\documentclass{article}

\PassOptionsToPackage{numbers, compress}{natbib}
\usepackage[preprint]{neurips_2026}

\usepackage[utf8]{inputenc} 
\usepackage[T1]{fontenc}    
\usepackage{hyperref}       
\usepackage{url}            
\usepackage{booktabs}       
\usepackage{amsfonts}       
\usepackage{nicefrac}       
\usepackage{microtype}      
\usepackage{xcolor}         
\usepackage{amsmath}

\usepackage{dcolumn}
\usepackage{multirow}
\usepackage{graphicx}
\usepackage{subcaption}
\usepackage{amssymb}
\usepackage{mathtools}
\usepackage{amsthm}
\usepackage{enumitem}
\usepackage{bm}
\usepackage{tikz}
\usetikzlibrary{arrows.meta,shapes.arrows,positioning,calc,fit}
\usetikzlibrary{shapes.arrows}
\usetikzlibrary{fit,backgrounds}
\usepackage[capitalize,noabbrev]{cleveref}
\usepackage[textsize=tiny]{todonotes}

\usepackage[most]{tcolorbox}

\newtcolorbox{contributionbox}{
  enhanced,
  breakable,
  colback=black!4,
  colframe=black!15,
  boxrule=0.4pt,
  arc=1mm,
  left=5pt,
  right=5pt,
  top=5pt,
  bottom=3pt,
  before skip=6pt,
  after skip=6pt
}

\newcommand{\metricup}{%
  \ensuremath{\mkern2mu{\color{black!40}\scriptstyle\uparrow}}}
\newcommand{\metricdown}{%
  \ensuremath{\mkern2mu{\color{black!40}\scriptstyle\downarrow}}}
\newcommand{\metrictoone}{%
  \ensuremath{\mkern2mu{\color{black!40}\scriptstyle\rightarrow 1}}}

\title{PerturbPFN: Probing the Limits of Synthetic Priors in Drug Perturbation Modelling}

\author{%
  Yuche Gao$^{1}$\thanks{Correspondence to \texttt{yg473@cam.ac.uk} and \texttt{sguo26v@gmail.com}.} \quad
  Jos\'e Miguel Hern\'andez-Lobato$^{1}$ \quad
  Siyuan Guo$^{2,1*}$ \\[0.6em]
  $^{1}$University of Cambridge, Cambridge, United Kingdom \\
  $^{2}$Prior Labs, Freiburg, Germany
}

\begin{document}

\maketitle

\begin{abstract}
Predicting cellular responses to unseen chemical perturbations is challenging due to unknown targets and mechanisms, high-dimensional expression responses, and limited experimental coverage of the large small-molecule design space. We propose PerturbPFN, a PFN-style amortized model for unknown-target perturbation prediction under a hierarchical synthetic structural prior. Instead of directly regressing high-dimensional expression responses, PerturbPFN infers a latent system graph, sparse atomic intervention targets, and intervention strengths, then propagates their effects through an SCM decoder. The model is trained entirely on prior-predictive synthetic episodes generated from biologically motivated graph and expression simulators, enabling structured in-context learning without test-time gradient updates. We evaluate PerturbPFN on both real single-cell perturbation data and synthetic benchmarks, covering effect prediction, target identification, and regulatory structure discovery. Our results show that PerturbPFN offers a complementary trade-off to specialized baselines, achieving competitive perturbation prediction with low inference cost while exposing interpretable intermediate estimates of targets, strengths, and system structure.
\end{abstract}

\section{Introduction}
Predicting how complex biological systems respond to novel interventions is a central challenge in drug discovery. The design of clinical therapeutics often relies on chemical perturbations  \citep{sadybekov2023computational}. Small-molecule drugs introduce a significantly more complex intervention modality than discrete genetic edits, such as CRISPR knockouts \citep{maathuis2010predicting}. For example, the perturbation target, pathway, and effective mechanism may be unknown, context-dependent, or only partially captured by perturbation covariates. In this regime, effect prediction requires not only modelling the post-perturbation distribution, but also inferring latent atomic targets and mechanisms from observed response data and perturbation features \citep{mooij2020joint,schneidergenerative}. Further, because the theoretical design space of these small molecules is vastly larger than what can be physically screened in a laboratory, interpolating between known drugs is insufficient. Predictive models must learn to generalize to unseen perturbations \citep{tejada2025causal}.

Existing deep learning approaches for perturbation prediction generally fall into three distinct paradigms. First, unstructured \textit{black-box} models directly learn the statistical distribution shifts induced by a perturbation \citep{hetzel2022predicting,bunne2023learning,lotfollahi2023predicting}, performing inference within a learned latent space. Second, structured \textit{mechanistic} models frame perturbation responses as explicit interventions on an underlying—often causal—representation of the system \citep{parascandolo2018learning, gonzalez2025combinatorial,roohani2024predicting}, offering advantages in interpretability. Finally, recent foundational model developments seek to learn robust representation spaces by pre-training on massive scales of real-world experimental data, utilizing broad biological priors as seen in scGPT \citep{cui2023scGPT}, GenePT \citep{Chen2023GenePT}, and LPM \citep{miladinovic2025silico}. 

However, current structured approaches often rely on instance-specific optimization or dataset-specific fitting, which limits their ability to amortize knowledge across heterogeneous contexts such as differing cell lines, experimental batches, or diverse perturbation panels. To overcome these limitations, we draw inspiration from Prior-Data Fitted Networks (PFNs), which have demonstrated that pre-training on large-scale synthetic tasks can amortize the inference process \citep{mullertransformers,hollmann2025accurate,robertson2025pfn,qu2025tabicl,grinsztajn2025tabpfn}. This enables ``in-context" prediction on novel datasets without the need for task-specific gradient updates.

\begin{figure*}[t]
  \centering
    \resizebox{\linewidth}{!}{%
        \input{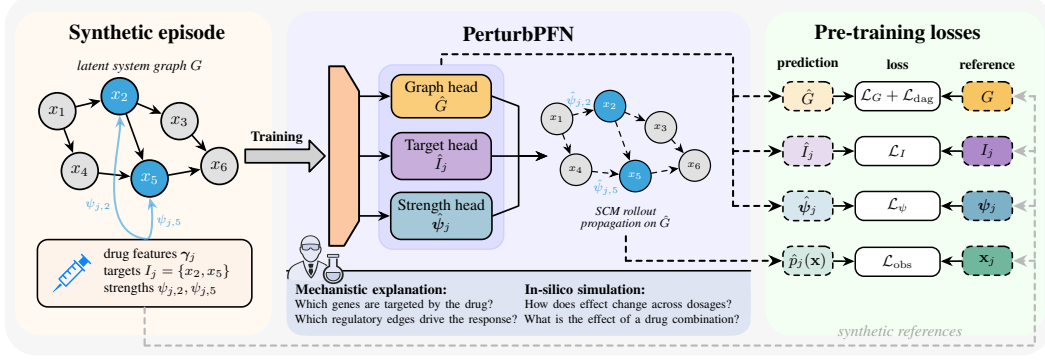}
    }
  \caption{PerturbPFN pretraining pipeline. Synthetic episodes provide latent references for the system graph \(G\), perturbation targets \(I_j\), intervention strengths \(\psi_j\), and outcomes \(\mathbf{x}_j\). The model predicts a graph and query-specific intervention variables, propagates them through an SCM decoder to predict perturbation effects, and is trained with multi-task losses on all synthetic references.}
  \label{fig:pipeline}
\end{figure*}

\begin{contributionbox}
\noindent\textbf{Our contributions.}
\begin{itemize}[leftmargin=*, itemsep=2pt, topsep=3pt]
    \item We introduce \textit{PerturbPFN}, a PFN-style framework for perturbation effect prediction under a hierarchical synthetic SCM prior, using biologically tailored generators to provide supervision for latent graphs, sparse intervention targets, and low-dimensional strength that are typically unavailable in empirical biological data, then propagates their effects through an inferred SCM decoder instead of directly regressing high-dimensional responses.

    \item PerturbPFN provides competitive effect prediction on real single-cell perturbation data \citep{srivatsan2020massively} through a single in-context forward pass, without dataset-specific optimization at test time.

    \item We evaluate latent graph, target, and strength inference on held-out synthetic episodes, and assess the transferability of the learned gene regulatory structure on external benchmarks \citep{nourisa2025genernib}.
\end{itemize}
\end{contributionbox}

\section{Methodology}
\label{sec:method}

This section introduces the SCM background, defines the episodic perturbation prediction setting and latent SCM factorization used by PerturbPFN, and describes the model architecture, training objective, and inference procedure.

\subsection{SCM Background and Shift Interventions}
\label{sec:scm-background}

A structural causal model (SCM) \(\mathcal{M}=\{G,\bm{\theta}\}\) represents a system by a directed acyclic graph (DAG) with adjacency matrix \(G\in\{0,1\}^{p\times p}\) over variables \(\mathbf{x}=(x_1,\ldots,x_p)\) and a collection of local mechanisms parameterized by \(\bm{\theta}\). The induced observational distribution factorizes as
\begin{equation}
p(\mathbf{x};G,\bm{\theta})
=
\prod_{i=1}^{p}
p_i(x_i\mid \mathbf{x}_{G_i};\bm{\theta}_i),
\end{equation}
where \(G_i\) denotes the parent set of variable \(i\). This factorization is useful because modifying one local mechanism can change the distribution of its target variable and propagate to downstream descendants through the graph.

An intervention on an SCM modifies a subset of local mechanisms. A general shift intervention changes the targeted mechanisms by applying a local shift or modulation,
\begin{equation}
p_i(x_i\mid \mathbf{x}_{G_i};\bm{\theta}_i)
\quad\longrightarrow\quad
\tilde p_i(x_i\mid \mathbf{x}_{G_i};\bm{\theta}_i,\psi_i),
\quad i\in I,
\end{equation}
while leaving non-targeted mechanisms unchanged. Unlike hard interventions, which set variables externally and remove their dependence on parents, shift interventions retain parent-dependent regulation. In this work, we use additive mechanism shifts in the latent SCM. Writing a local mechanism as
\begin{equation}
x_i = f_i(\mathbf{x}_{G_i};\bm{\theta}_i)+\epsilon_i,
\end{equation}
a targeted intervention with strength \(\psi_i\) changes it to
\begin{equation}
x_i = f_i(\mathbf{x}_{G_i};\bm{\theta}_i)+\psi_i+\epsilon_i,
\quad i\in I,
\end{equation}
with \(\psi_i=0\) for non-targeted variables. The post-intervention distribution is obtained by rolling out the modified SCM in topological order.

\subsection{Problem Setup}
We study episodic perturbation prediction. Each episode represents one system observed under multiple perturbation regimes, such as a cell line exposed to different drugs. The system contains \(p\) measured variables. Each observation is written as \((\bm{\gamma}_i,\mathbf{x}_i)\), where \(\mathbf{x}_i\in\mathbb{R}^p\) is the measured system state and \(\bm{\gamma}_i\in\mathbb{R}^d\) is a regime descriptor that parameterizes experimental conditions. For example, \(\bm{\gamma}_i\) may encode the chemical embedding of a specific treatment, a continuous scalar for dosage levels, or a categorical indicator for control treatments. Given context observations
\(
\mathcal{D}_{\mathrm{ctx}}=\{(\bm{\gamma}_i,\mathbf{x}_i)\}_{i=1}^{n_{\mathrm{ctx}}},
\)
and a query regime \(\bm{\gamma}_j\), the desired prediction is the posterior predictive distribution (PPD)
\begin{equation}
p(\mathbf{x}_j\mid \bm{\gamma}_j,\mathcal{D}_{\mathrm{ctx}})
=
\int
p(\mathbf{x}_j\mid \bm{\gamma}_j,\mathbf{t})\,
p(\mathbf{t}\mid \mathcal{D}_{\mathrm{ctx}})
\,d\mathbf{t},
\end{equation}
where \(\mathbf{t}\) denotes latent task variables, including the system structure and perturbation-response mechanisms that map each regime descriptor \(\bm{\gamma}_j\) to unobserved atomic targets and intervention strengths. PerturbPFN approximates this target with point estimates of the latent graph, targets, and strengths.

\paragraph{Latent intervention model.}

We assume that each episode is governed by a latent causal generative process
\(\mathcal{M}=\{G,\bm{\theta}\}\), where \(G\in\{0,1\}^{p\times p}\) is the adjacency matrix of a directed acyclic graph (DAG) over the measured variables and \(\bm{\theta}\) parameterizes the local mechanisms. A perturbation regime \(\bm{\gamma}\) does not directly determine the full high-dimensional response. Instead, it induces a latent atomic intervention \(\mathcal{I}=\{I,\bm{\psi}\}\), where \(I\in\{0,1\}^{p}\) denotes sparse intervention targets and \(\bm{\psi}\) denotes their strengths. We factorize this intervention model as
\begin{equation}
    p(\mathcal{I}\mid \bm{\gamma}, \bm{\phi})
    =
    p\bigl(I;\, g_{\bm{\phi}}(\bm{\gamma})\bigr)p\bigl(\bm{\psi};\, h_{\bm{\phi}}(I,\bm{\gamma})\bigr),
    \label{eq:intervention-model}
\end{equation}
where \(g_{\bm{\phi}}\) and \(h_{\bm{\phi}}\) are descriptor-to-intervention maps parameterized by \(\bm{\phi}\): \(g_{\bm{\phi}}\) maps perturbation descriptors to target probabilities, and \(h_{\bm{\phi}}\) maps targets and descriptors to intervention-strength parameters. In PerturbPFN, these descriptor-to-intervention maps are amortized by MLPs that take context-conditioned node and query representations as input. Given \(\mathcal{I}\), the perturbation modifies only the targeted mechanisms through shift interventions. PerturbPFN uses this decomposition as a structural inductive bias: it predicts point estimates of \(G\), \(\mathbf{I}\), and \(\bm{\psi}\), and decodes the response through the corresponding intervened causal generative process.

\subsection{Hierarchical Synthetic Prior}
PerturbPFN is trained on prior-predictive episodes from a hierarchical synthetic SCM prior tailored to single-cell perturbation data. At the episode level, the prior samples a gene-regulatory network (GRN) using a generator inspired by \citet{aguirre2025gene}, augmented with module structure to induce heterogeneous, scale-free topology with stronger within-module regulation \cite{bartlett2026inferring}. At the perturbation level, descriptors \(\bm{\gamma}\) generate sparse atomic targets from a truncated geometric distribution target-count prior, with similar descriptors tending to produce related targets and effect strengths. Interventions apply signed additive shifts whose magnitudes follow Hill-type dose--response functions, and nonlinear local mechanisms propagate these shifts to induce downstream changes in non-target variables. Observations are then sampled from a zero-inflated log-normal expression model with SERGIO noise injection \cite{dibaeinia2020sergio}. For synthetic counterfactual supervision, we reuse exogenous noise across factual and intervened rollouts so that paired differences primarily reflect intervention-induced changes rather than independent sampling noise \cite{sextro2026mappfn}. The prior supplies synthetic labels for graphs, targets, strengths, and outcomes; no real perturbation responses are used for training. Further details on synthetic data generation are provided in Appendix~\ref{app:synthetic}.

\subsection{PerturbPFN Architecture}
\label{sec:architecture}

PerturbPFN implements the latent factorization above with a transformer-based amortized predictor.
Given context perturbation--response pairs and a query regime, the model first infers intervention variables for the context regimes and an episode-level system graph. It then conditions the query on the inferred context interventions, predicts query-specific targets and strengths, and rolls out an intervened SCM decoder.

\paragraph{Context intervention memory and structure branch.}
The shared context encoder processes perturbation--response pairs
\((\bm{\gamma}_i,\mathbf{x}_i)\) and produces node-level representations, regime summaries \(z_i^{\mathrm{ctx}}\), and a global episode summary.
The target and strength heads are first applied to every context regime to obtain
\(\hat{\mathbf I}_i\) and \(\hat{\bm{\psi}}_i\). These predictions, together with the regime representation, descriptor, and observed response shift
\(\Delta\mathbf{x}_i\), are encoded into one intervention-memory token per context regime:
\begin{equation}
m_i
=
E_{\mathrm{mem}}
\!\left(
z_i^{\mathrm{ctx}},
\bm{\gamma}_i,
\hat{\mathbf I}_i,
\hat{\bm{\psi}}_i,
\Delta\mathbf{x}_i
\right),
\quad
\tilde q_j
=
q_j
+
\operatorname{MHA}
\!\left(
q_j,
\{m_i\}_{i\in\mathrm{ctx}},
\{m_i\}_{i\in\mathrm{ctx}}
\right).
\end{equation}
Here, \(q_j\) combines the query descriptor with the global episode representation, while \(\tilde q_j\) is passed to the query target, strength, and mechanism-modulation heads.

The structure branch uses a set-to-graph encoder \citep{dhirmeta} to combine context response statistics with the inferred targets and strengths:
\begin{equation}
H^G
=
E^G_\eta
\!\left(
\mathcal{D}_{\mathrm{ctx}},
\hat{\mathbf I}_{\mathrm{ctx}},
\hat{\bm{\psi}}_{\mathrm{ctx}}
\right).
\end{equation}
Across context regimes, associations between interventions on node \(u\) and response shifts at node \(v\) provide evidence for the directed edge \(u\to v\). The graph is shared across query regimes within an episode and is therefore not conditioned on a particular query. During structure pretraining, graph evidence uses the synthetic ground-truth context interventions; joint training and deployment instead use the inferred context targets and strengths.

\paragraph{Structure head and DAG projection.}
For each ordered pair \(u\neq v\), the structure head uses a directed edge scorer augmented with context-level intervention--response evidence:
\begin{equation}
\ell_{uv}
= f_{\mathrm{edge}}
\!\left(h^G_u,h^G_v;\mathcal{C}_{\mathrm{ctx}}\right),\quad
P_{uv}=
\sigma(\ell_{uv}),
\quad P_{uu}=0,
\end{equation}
where \(\mathcal{C}_{\mathrm{ctx}}\) summarizes pairwise context representations and the relationships between inferred interventions and response shifts. In our implementation, \(f_{\mathrm{edge}}\) combines a biaffine source--destination scorer with learned context-pair and intervention--response features.

To encourage acyclicity, we apply the NOTEARS-style penalty \citep{zheng2018dags}
\begin{equation}
\mathcal{L}_{\mathrm{dag}}
=
\operatorname{tr}\!\left(\exp(P\odot P / p)\right)-p .
\end{equation}
Rather than relying on a single fixed probability threshold, hard graph extraction calibrates an episode-specific threshold to a target edge density estimated from the synthetic prior:
\begin{equation}
\hat G
=
\Pi_{\mathrm{DAG}}(P;\rho_G).
\end{equation}
Candidate edges are sorted by score and greedily added until the target density is reached, while edges that create directed cycles are skipped. During synthetic training, the calibration can use the known graph density; at inference it uses the fixed prior-derived density. The resulting graph is therefore sparse and acyclic.

\paragraph{Intervention heads and SCM rollout.}
Conditioned on the attended query representation, the target head predicts node-wise probabilities and constructs a sparse straight-through top-\(k\) target gate \(\hat{\mathbf I}_j\). The strength head predicts a categorical bar distribution over intervention strengths and uses its expectation as \(\hat{\bm{\psi}}_j\). A separate modulation head produces node- and query-specific scale, shift, and gate parameters for the SCM decoder.

Let \(z_u\) denote the generated latent log-state of variable \(u\). For each variable \(v\), the decoder represents every potential parent \(u\) in a dedicated input slot. The hard- and soft-graph parent inputs are
\begin{equation}
\mathbf m_v^{\mathrm{hard}}
=
\bigl[\hat G_{uv}z_u\bigr]_{u=1}^{p},
\qquad
\mathbf m_v^{\mathrm{soft}}
=
\bigl[\hat G_{uv}P_{uv}z_u\bigr]_{u=1}^{p}.
\end{equation}
A shared nonlinear decoder, modulated by the node- and query-specific parameters, generates variables in topological order. The intervention enters once as an additive shift to the targeted latent mechanism,
\begin{equation}
z_v
=
f_\theta
\!\left(
\mathbf m_v,\mathbf{q}_{j,v}
\right)
+
\hat I_{j,v}\hat\psi_{j,v},
\end{equation}
and the latent location \(z_v\), rather than the zero-inflated observation mean, is propagated to descendants. The decoder outputs the location, variance, and zero probability of a zero-inflated log-normal predictive distribution. Hard projected graphs are used at test time.

\subsection{Training Curriculum and Inference}
\label{sec:training-inference}

PerturbPFN is trained entirely on prior-predictive synthetic episodes, where the latent graph \(G\), intervention targets \(\mathbf{I}_j\), strengths \(\bm{\psi}_j\), and outcomes \(\mathbf{x}_j\) are known by construction. As shown in Figure~\ref{fig:pipeline}, the training objective is
\begin{equation}
\mathcal{L}
=
\lambda_G\mathcal{L}_G
+
\lambda_I\mathcal{L}_I
+
\lambda_\psi\mathcal{L}_\psi
+
\mathcal{L}_{\mathrm{obs}}
+
\mathcal{A}_{\mathrm{dag}} .
\end{equation}
Here, \(\mathcal{L}_G\) combines an edge-level graph loss with density regularization. \(\mathcal{L}_I\) aggregates weighted focal binary-cross-entropy losses for query and context targets. \(\mathcal{L}_\psi\) uses a target-aware bar-distribution likelihood for query and context strengths. \(\mathcal{L}_{\mathrm{obs}}\) is the negative log-likelihood of query responses under the zero-inflated log-normal decoder distribution, while \(\mathcal{A}_{\mathrm{dag}}\) is an adaptive augmented-Lagrangian penalty based on the NOTEARS acyclicity measure.

Training uses two stages for the checkpoint reported in Section~\ref{sec:experiments}. We first pretrain the structure branch using synthetic graph labels, with ground-truth context interventions used to construct structure evidence. We then jointly train the full model while rolling out the decoder on the ground-truth synthetic graph and supervising query and context intervention variables together with query outcomes. At test time, PerturbPFN performs one model forward pass without parameter updates. Continuous edge scores are converted into a hard graph using prior-density calibration followed by greedy DAG projection, and the decoder is rolled out on this hard graph. Further architecture and training details are provided in Appendix~\ref{app:architecture}.

\section{Related Work}

\paragraph{Perturbation response prediction.}
Many single-cell perturbation models treat the task as supervised counterfactual response prediction. Autoencoder and latent-variable methods such as scGen \cite{lotfollahi2019scgen}, CPA \cite{lotfollahi2023predicting}, Biolord \cite{piran2024disentanglement}, and SAMS-VAE \cite{bereket2023modelling} learn perturbation or attribute representations that are recombined with cell-state embeddings to decode post-perturbation expression profiles, with ChemCPA additionally using chemical descriptors for unseen-drug prediction \cite{hetzel2022predicting}. Optimal-transport methods such as CellOT \cite{bunne2023learning} and CondOT \cite{bunne2022supervised} instead learn maps between unpaired control and treated-cell distributions, optionally conditioned on perturbation context. Knowledge-informed predictors such as GEARS incorporate gene--gene graphs with graph neural networks for genetic perturbation prediction \cite{roohani2024predicting}. These methods can be effective, but they typically predict responses end-to-end through latent spaces, transport maps, or external knowledge graphs, rather than jointly inferring unknown atomic targets, intervention strengths, and propagation through an inferred SCM.

\paragraph{Structured perturbation models.}
A complementary line of work represents perturbations as interventions on structured or causal models, aiming to recover mechanisms rather than only predict expression shifts. Classical multi-environment causal discovery methods, including JCI-PC \cite{mooij2020joint}, UT-IGSP \cite{squires2020permutation}, and BaCaDi \citep{hagele2023bacadi}, infer causal graphs and, in some cases, unknown intervention targets from observational and interventional data. More recent perturbation models make this link explicit: GIM learns a generative map from perturbation descriptors to atomic interventions in a causal model \cite{schneidergenerative}, while SCCVAE combines a variational autoencoder with a learned regulatory network and shift interventions for genetic perturbation response prediction \citep{liu2026learning}. Related inverse-design methods such as PDGrapher use causally inspired graph neural networks to predict therapeutic target sets that would move a diseased state toward a desired treated state \citep{gonzalez2025combinatorial}. These approaches provide stronger mechanistic structure than end-to-end predictors, but their reliance on instance-specific fitting, iterative inference, or predefined proxy graphs increases computational cost and limits amortized transfer across new perturbation contexts.

\paragraph{Prior-data fitted networks and single-cell foundation models.}
Prior-data fitted networks (PFNs) show that transformers pretrained on synthetic tasks can amortize Bayesian-style inference and make predictions on new datasets by conditioning on examples in context, motivating recent tabular foundation models such as TabPFN and TabICL \citep{mullertransformers,hollmann2025accurate,qu2026tabiclv2}. This paradigm has recently been extended to causal inference, including amortized causal effect estimation \citep{robertson2025pfn,balazadeh2025causalpfn}, interventional prediction under graph uncertainty \cite{dhir2026estimating}, and general PFN frameworks for causal reasoning \citep{ma2025foundation}; MapPFN is especially related, learning causal perturbation maps for biological response prediction in context \citep{sextro2026mappfn}. In parallel, single-cell foundation models such as scGPT \cite{cui2023scGPT}, Geneformer \citep{theodoris2023transfer}, and GenePT \cite{Chen2023GenePT} learn transferable gene or cell representations from transcriptomic corpora or biomedical text, while perturbation-oriented models such as LPM \cite{miladinovic2025silico} and Stack \citep{dong2026stack} use symbolic perturbation--readout--context tuples or in-context sets of cells for response prediction. PerturbPFN combines these directions: it uses PFN-style synthetic-prior pretraining, but specializes the prior to structured perturbation mechanisms, explicitly predicting latent graphs, unknown targets, and intervention strengths before decoding the perturbation response.

\section{Experiments}
\label{sec:experiments}

We evaluate PerturbPFN along three axes: in-prior latent recovery, real perturbation effect prediction, and GRN structure discovery. Together, these tests assess whether a single synthetic-only checkpoint learns a non-collapsed graph--target--strength bottleneck, predicts held-out perturbation responses, and transfers useful regulatory signal to external biological data. All evaluations use \textbf{the same pretrained checkpoint} without task-specific gradient updates.

\begin{figure*}[t]
  \centering
  \includegraphics[width=\textwidth]{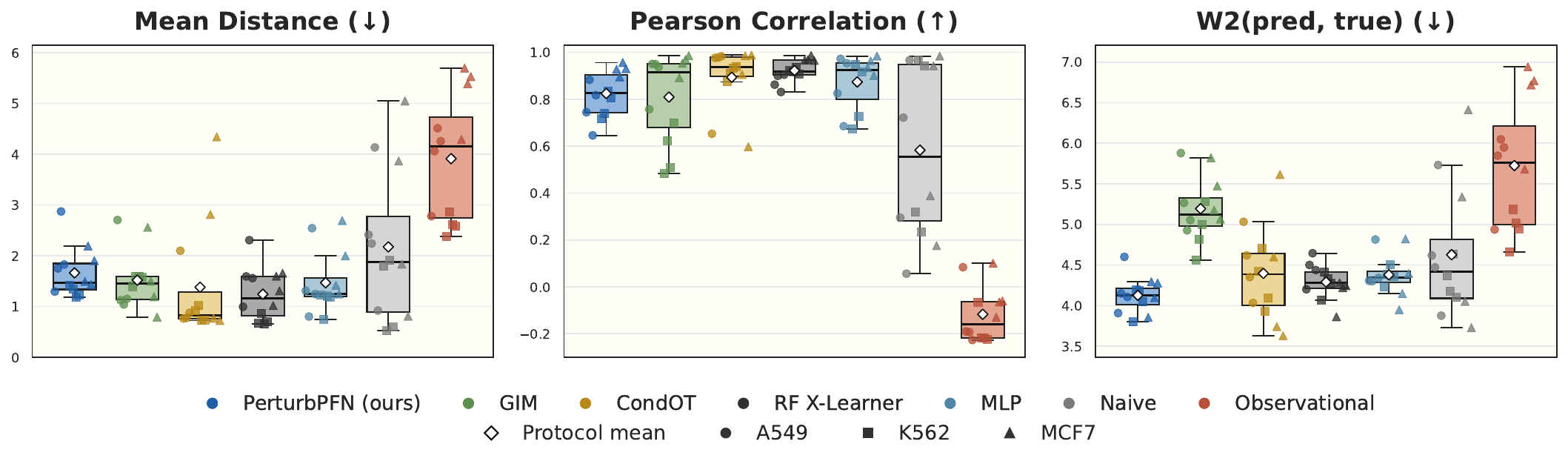}
  \caption{Raw effect-prediction performance on the 12-protocol highest-dosage holdout benchmark from \citet{srivatsan2020massively}. Each point represents one held-out cell line--drug protocol, and boxplots summarize performance across protocols. PerturbPFN achieves competitive performance across all three metrics and the strongest distributional accuracy measured by \(W_2\).}
  \label{fig:4metrics}
\end{figure*}

\subsection{In-Prior Latent Recovery}
\label{sec:in-prior}

\paragraph{Protocol.}
We first evaluate whether PerturbPFN recovers the latent variables represented by its structured bottleneck on \(1{,}024\) fresh episodes drawn from the synthetic prior. Because the latent graph, intervention targets, strengths, and response distributions are known by construction, this evaluation isolates in-prior amortized inference before considering transfer to real biological data.

\paragraph{Metrics.}
For graph recovery, AUROC and AUPRC evaluate the ranking of all directed non-self edges. We additionally report F1 after calibrating the extraction threshold to the ground-truth graph density and greedily projecting the selected edges to a DAG. This density-matched F1 measures whether the highest-ranked edges correspond to the true graph, while avoiding confounding from threshold calibration. Target AUROC and AUPRC evaluate node-wise target scores, and target F1 thresholds the predicted probabilities at \(0.5\). The target-strength ratio compares the mean absolute predicted and true strengths on targeted nodes. Finally, response-shift cosine similarity measures directional agreement between predicted and true mean intervention effects, while observation NLL evaluates the full zero-inflated log-normal predictive distribution. See Appendix~\ref{app:in-prior} for details.

\begin{table*}[t]
\centering
\caption{In-prior latent recovery over \(128\) held-out synthetic batches. Graph F1 uses the top-scoring acyclic edges at the ground-truth graph density.}
\label{tab:in_prior_sanity}
\small
\begin{tabular}{lclclc}
\toprule
\multicolumn{2}{c}{Graph} &
\multicolumn{2}{c}{Target} &
\multicolumn{2}{c}{Strength / Response} \\
\cmidrule(lr){1-2}
\cmidrule(lr){3-4}
\cmidrule(lr){5-6}
AUROC\metricup & 0.8133 &
AUROC\metricup & 0.9330 &
\(\lvert\psi\rvert\) ratio\metrictoone & 0.9386 \\
AUPRC\metricup & 0.2980 &
AUPRC\metricup & 0.7860 &
\(\Delta\) cosine\metricup & 0.7437 \\
Proj. F1\metricup & 0.3261 &
F1\metricup & 0.7649 &
Obs. NLL\metricdown & -0.1375 \\
\bottomrule
\end{tabular}
\end{table*}

\paragraph{Results.}
Table~\ref{tab:in_prior_sanity} shows that all components of the latent bottleneck remain non-collapsed. Graph scores achieve an AUROC of \(0.8133\) and an AUPRC of \(0.2980\), substantially above the underlying edge prevalence of \(0.0363\). Target inference is even stronger, reaching an AUROC of \(0.9330\), an AUPRC of \(0.7860\), and a thresholded F1 of \(0.7649\). The straight-through top-\(k\) gate used by the decoder obtains a lower F1 of \(0.6287\), reflecting the additional sparsity constraint imposed during rollout. Predicted target-node strengths recover the overall intervention scale, with a mean absolute-strength ratio of \(0.9386\). The response-shift cosine similarity of \(0.7437\) further shows that the decoder recovers much of the intervention direction. These results establish that PerturbPFN learns a functional graph--target--strength bottleneck under its training prior; transfer to real biological data is evaluated below.

\subsection{Perturbation Effect Prediction}
\label{sec:effect-prediction}

\paragraph{Protocol.}
We evaluate perturbation-response prediction on the Sci-Plex single-cell drug perturbation dataset \citep{srivatsan2020massively}, using the highest-dosage holdout protocol of \citet{schneidergenerative}. The benchmark contains \(12\) cell line--drug protocols formed from three cell lines and four drugs. For each protocol, the highest-dose response is held out, while the model receives lower doses of the same drug and all observed doses of the remaining drugs. It must predict the held-out cell distribution without observing drug targets or updating model parameters. Full construction details are provided in Appendix~\ref{app:effect-prediction}.

\paragraph{Metrics.}
We report three complementary metrics. Mean distance,
\(\mathrm{MD}=\lVert\hat{\bm\mu}_{\mathrm{int}}-\bm\mu_{\mathrm{int}}\rVert_2\),
measures error between predicted and true mean-response vectors. Pearson correlation \(r\) measures their gene-wise agreement. \(W_2(\hat X_{\mathrm{int}},X_{\mathrm{int}})\) compares the full predicted and observed cell distributions; we approximate it by a debiased Sinkhorn estimator. This is particularly important for perturbation modelling, where a model may obtain reasonable correlation while substantially under- or over-estimating the magnitude of the response. 

For each protocol and metric, we orient performance so that higher is better and min--max normalize scores across models. We then average over the \(12\) protocols to obtain each metric column and average the three columns for the overall ranking in Table~\ref{tab:runtime-per-protocol}. Because min--max normalization is relative to the models being compared, Figure~\ref{fig:4metrics} additionally reports the raw protocol-level distributions. Table~\ref{tab:runtime-per-protocol} also reports runtime per protocol. For PerturbPFN, the ICL-only runtime measures application of the pretrained checkpoint, whereas the amortized runtime additionally allocates its one-time pretraining cost across the \(12\) protocols. Runtimes were measured on the same compute worker equipped with a single NVIDIA RTX 4090; further compute details are provided in Appendix~\ref{app:architecture}.

\paragraph{Baselines.}
We compare against GIM \citep{schneidergenerative}, CondOT \citep{bunne2022supervised}, a multilayer perceptron (MLP) and a random-forest X-Learner (RF X-Learner) \citep{kunzel2019metalearners}. We also include a Naive baseline that uses the nearest observed lower dose of the held-out drug and an Observational baseline that predicts the control distribution.

\begin{table*}[t]
\centering
\caption{Runtime per protocol and normalized Sci-Plex effect-prediction scores. Times are mean \(\pm\) standard deviation. PerturbPFN (amortized) allocates its one-time pretraining cost across the \(12\) protocols, whereas PerturbPFN (ICL only) reports application of the pretrained model.}
\label{tab:runtime-per-protocol}
\small
\begin{tabular}{lD{+}{\,\pm\,}{-1}cccc}
\toprule
Model & \multicolumn{1}{c}{Time (s) / protocol} &
\(\hat{s}_{\mathrm{MD}}\)\metricup &
\(\hat{s}_{r}\)\metricup &
\(\hat{s}_{W_2}\)\metricup &
Avg. \(\hat{s}\) / rank \\
\midrule
RF X-Learner
& 1.83+0.32
& \textbf{0.8793}
& \textbf{0.9696}
& 0.8122
& \textbf{0.8870} (\#1) \\

CondOT
& 7266.4+585.9
& 0.8014
& 0.9323
& 0.7401
& 0.8246 (\#3) \\

MLP
& 0.82+0.25
& 0.7823
& 0.9216
& 0.7582
& 0.8207 (\#4) \\

GIM
& 4495.6+116.9
& 0.6909
& 0.8611
& 0.2934
& 0.6151 (\#5) \\

Naive
& \multicolumn{1}{c}{\quad /}
& 0.5435
& 0.6293
& 0.6299
& 0.6009 (\#6) \\

Observational
& \multicolumn{1}{c}{\quad /}
& 0.0000
& 0.0000
& 0.0698
& 0.0233 (\#7) \\

\midrule
PerturbPFN (amortized)
& 432.8 + 0.34
& \multirow{2}{*}{0.7186}
& \multirow{2}{*}{0.8813}
& \multirow{2}{*}{\textbf{0.9183}}
& \multirow{2}{*}{0.8394 (\#2)} \\

PerturbPFN (ICL only)
& 1.11 + 0.34
& & & & \\
\bottomrule
\end{tabular}%
\end{table*}

\paragraph{Results.}
PerturbPFN achieves the best normalized \(W_2\) score and ranks second overall, behind RF X-Learner. Once pretrained, it requires \(1.11\) seconds per protocol, faster than all model-based baselines except MLP; even after amortizing pretraining, its \(432.8\)-second cost remains substantially below GIM and CondOT. Although PerturbPFN does not uniformly dominate the real-data baselines, it offers a favorable trade-off between predictive performance, structured interpretability, and test-time cost: Among the evaluated baselines, GIM is the only method that provides comparable latent estimates of graph structure, intervention targets, and strengths. However, PerturbPFN is substantially faster than GIM and achieves markedly stronger effect prediction.

\subsection{GRN Structure Discovery}
\label{sec:grn-structure-discovery}

\paragraph{Protocol.}
We evaluate whether the learned graph head transfers to external GRN recovery using the official GeneRNIB 300BCG benchmark \citep{nourisa2025genernib}. Following the official split, methods infer a weighted, directed TF--gene network from baseline expression profiles collected from \(38\) donors. Held-out three-month profiles and external regulatory references are used only for evaluation. Each method outputs a ranked list of at most \(50{,}000\) directed TF--gene links, allowing the benchmark to evaluate both the predictive and biological utility of the inferred network.

PerturbPFN is trained to infer an intervention-induced causal dependency graph over measured variables, which is related to but not identical to a curated GRN. GeneRNIB evaluates agreement with TF--target references, pathway activity, and predictive utility, whereas PerturbPFN's graph is optimized to propagate intervention effects under a synthetic SCM prior. The benchmark therefore evaluates whether its learned causal structure transfers to biologically meaningful regulatory relationships.

\paragraph{Metrics and baselines.}
Because a complete ground-truth human GRN is unavailable, GeneRNIB uses complementary biological and predictive metrics. GS-F1 compares pathways enriched among predicted TF targets with pathways active in the held-out expression data, averaged across multiple gene-set collections. TFB-F1 measures normalized agreement and coverage against TF-binding references from UniBind, ChIP-Atlas, and ReMap. R-Precision and R-Recall evaluate how well TFs selected from the inferred network predict target-gene expression under conservative and broader regulator budgets, respectively. VC evaluates whether a GRN-constrained virtual-cell model can predict held-out perturbed expression. Higher values are better for all metrics; we omit replicate consistency for space.

We compare against Pearson correlation and partial Pearson correlation (PPCOR) \citep{kim2015ppcor}, the tree-based GRNBoost2 method \citep{moerman2019grnboost2}, the expression-based PORTIA method \citep{passemiers2022fast}, and SCENIC \citep{aibar2017scenic}, which combines co-expression inference with motif-based regulon pruning. All methods are evaluated using the same official GeneRNIB pipeline and link budget.

\begin{table}[t]
\centering
\caption{GeneRNIB 300BCG GRN recovery. Higher is better for all metrics.}
\label{tab:genernib_300bcg}
\small
\begin{tabular}{lccccc}
\toprule
Model & GS-F1 & TFB-F1 & R-Prec. & R-Recall & VC \\
\midrule
Pearson corr. & 0.4693 & 0.1102 & \textbf{0.6154} & 0.2742 & \textbf{0.1882} \\
PPCOR         & 0.0982 & 0.0000 & 0.1742 & 0.1920 & 0.0660 \\
GRNBoost2     & 0.5566 & 0.1129 & 0.6082 & \textbf{0.3752} & 0.1576 \\
PORTIA        & 0.6188 & 0.1014 & 0.3970 & 0.2668 & 0.1169 \\
SCENIC        & 0.4813 & \textbf{0.1180} & 0.3802 & 0.2747 & 0.0672 \\
\midrule
PerturbPFN    & \textbf{0.6410} & 0.1146 & 0.2250 & 0.1977 & 0.0731 \\
\bottomrule
\end{tabular}
\end{table}

\paragraph{Results.}
PerturbPFN achieves the best GS-F1 on 300BCG and is competitive on TFB-F1. This suggests that the learned graph head captures transferable regulatory signal despite being trained only on synthetic prior-predictive episodes. However, the gains are not uniform: dedicated GRN baselines remain stronger on R-Precision, R-Recall, and VC. This is expected because PerturbPFN is not optimized to reproduce a static GRN annotation target; it learns a causal propagation graph for perturbation response modelling.

Compared with conventional causal discovery methods \citep{guo2023causal}, such as BaCaDi* \citep{hagele2023bacadi} and JCI-PC \citep{mooij2020joint,spirtes2000causation}, PerturbPFN amortizes structural inference from synthetic biological priors rather than fitting each instance under relatively simple structural assumptions. Biological regulatory networks exhibit sparsity, modularity, degree heterogeneity, and other topological regularities captured by specialized GRN generators \citep{aguirre2025gene,dibaeinia2020sergio,barabasi1999emergence}. Training on these generators allows PerturbPFN to exploit simulator-defined biological priors, a key strength of PFNs \citep{muller2025position}. Consistent with this interpretation, PerturbPFN achieves edge AUROC \(0.9168\) and edge average precision \(0.7570\) in the auxiliary graph-ranking evaluation (Appendix~\ref{app:regulatory-structure}), although these metrics are not included in the official GeneRNIB score table.

\section{Discussion}
\label{sec:discussion}

PerturbPFN illustrates a broader route toward structured foundation models for biology: rather than scaling only with empirical observations, a model can be pretrained over mechanistic hypotheses sampled from a structured simulator. In this view, the transformer amortizes an inverse problem---inferring which latent graph, intervention targets, and mechanism changes could explain the observed context---and applies the resulting inference procedure to new perturbation tasks without gradient-based adaptation. The structured bottleneck therefore serves not only as an architectural constraint, but also as a way to expose inspectable hypotheses about how a perturbation acts and propagates through a system.

\paragraph{Limitations and Outlook.}
The meaning of PerturbPFN's latent variables is determined by its synthetic prior. Recovering them in-prior does not guarantee that they correspond to uniquely identifiable biological mechanisms, and a causal propagation graph need not coincide with a static curated GRN. The current model also represents graphs, targets, and strengths using point estimates, assumes a sparse acyclic system, and operates on gene panels much smaller than a complete transcriptome. Future work could employ richer and empirically calibrated simulators, quantify posterior uncertainty over latent mechanisms, support context-dependent or cyclic regulatory dynamics, and scale structured inference by composing predictions over local gene subsystems. These directions would strengthen the connection between efficient amortized prediction and biologically grounded mechanism discovery.

\begin{ack}
This work used resources provided by the Cambridge Service for Data-Driven Discovery (CSD3), operated by the University of Cambridge Research Computing Service (www.csd3.cam.ac.uk), provided by Dell EMC and Intel using Tier-2 funding from the Engineering and Physical Sciences Research Council (capital grant EP/T022159/1), and DiRAC funding from the Science and Technology Facilities Council (www.dirac.ac.uk). 
Additional experiments were run on GPU resources provided by the Computational and Biological Learning Lab (CBL) at the University of Cambridge. Y.G. gratefully acknowledges support from the Student Educational Development Award from St Edmund's College, University of Cambridge.
\end{ack}

\bibliographystyle{unsrtnat}
\bibliography{references}

@inproceedings{mullertransformers,
  title={Transformers Can Do Bayesian Inference},
  author={M{\"u}ller, Samuel and Hollmann, Noah and Arango, Sebastian Pineda and Grabocka, Josif and Hutter, Frank},
  booktitle={International Conference on Learning Representations},
  year={2022}
}

@article{hollmann2025accurate,
  title={Accurate predictions on small data with a tabular foundation model},
  author={Hollmann, Noah and M{\"u}ller, Samuel and Purucker, Lennart and Krishnakumar, Arjun and K{\"o}rfer, Max and Hoo, Shi Bin and Schirrmeister, Robin Tibor and Hutter, Frank},
  journal={Nature},
  volume={637},
  number={8045},
  pages={319--326},
  year={2025},
  publisher={Nature Publishing Group UK London}
}

@inproceedings{robertson2025pfn,
  title={Do-PFN: In-Context Learning for Causal Effect Estimation},
  author={Robertson, Jake and Reuter, Arik and Guo, Siyuan and Hollmann, Noah and Hutter, Frank and Sch{\"o}lkopf, Bernhard},
  booktitle={The Thirty-ninth Annual Conference on Neural Information Processing Systems},
  year={2025},
}

@article{bartlett2026inferring,
  title     = {Inferring gene-regulatory networks using epigenomic priors},
  author    = {Bartlett, Thomas E. and Li, Melodie and Song, Chenyu and Gao, Yuche and Huang, Qiulin},
  journal   = {iScience},
  volume    = {29},
  pages     = {115165},
  year      = {2026},
  publisher = {Elsevier},
  doi       = {10.1016/j.isci.2026.115165}
}

@article{sadybekov2023computational,
  title={Computational approaches streamlining drug discovery},
  author={Sadybekov, Anastasiia V and Katritch, Vsevolod},
  journal={Nature},
  volume={616},
  number={7958},
  pages={673--685},
  year={2023},
  publisher={Nature Publishing Group UK London}
}

@article{tejada2025causal,
  title={Causal machine learning for single-cell genomics},
  author={Tejada-Lapuerta, Alejandro and Bertin, Paul and Bauer, Stefan and Aliee, Hananeh and Bengio, Yoshua and Theis, Fabian J},
  journal={Nature Genetics},
  volume={57},
  number={4},
  pages={797--808},
  year={2025},
  publisher={Nature Publishing Group US New York}
}

@article{hetzel2022predicting,
  title={Predicting cellular responses to novel drug perturbations at a single-cell resolution},
  author={Hetzel, Leon and Boehm, Simon and Kilbertus, Niki and G{\"u}nnemann, Stephan and Theis, Fabian and others},
  journal={Advances in Neural Information Processing Systems},
  volume={35},
  pages={26711--26722},
  year={2022}
}

@article{lotfollahi2023predicting,
  title={Predicting cellular responses to complex perturbations in high-throughput screens},
  author={Lotfollahi, Mohammad and Klimovskaia Susmelj, Anna and De Donno, Carlo and Hetzel, Leon and Ji, Yuge and Ibarra, Ignacio L and Srivatsan, Sanjay R and Naghipourfar, Mohsen and Daza, Riza M and Martin, Beth and others},
  journal={Molecular systems biology},
  volume={19},
  number={6},
  pages={MSB202211517},
  year={2023},
  publisher={Springer}
}

@article{bunne2023learning,
  title={Learning single-cell perturbation responses using neural optimal transport},
  author={Bunne, Charlotte and Stark, Stefan G and Gut, Gabriele and Del Castillo, Jacobo Sarabia and Levesque, Mitch and Lehmann, Kjong-Van and Pelkmans, Lucas and Krause, Andreas and R{\"a}tsch, Gunnar},
  journal={Nature methods},
  volume={20},
  number={11},
  pages={1759--1768},
  year={2023},
  publisher={Nature Publishing Group US New York}
}

@inproceedings{parascandolo2018learning,
  title={Learning independent causal mechanisms},
  author={Parascandolo, Giambattista and Kilbertus, Niki and Rojas-Carulla, Mateo and Sch{\"o}lkopf, Bernhard},
  booktitle={International Conference on Machine Learning},
  pages={4036--4044},
  year={2018},
  organization={PMLR}
}

@article{gonzalez2025combinatorial,
  title={Combinatorial prediction of therapeutic perturbations using causally inspired neural networks},
  author={Gonzalez, Guadalupe and Lin, Xiang and Herath, Isuru and Veselkov, Kirill and Bronstein, Michael and Zitnik, Marinka},
  journal={Nature Biomedical Engineering},
  pages={1--18},
  year={2025},
  publisher={Nature Publishing Group UK London}
}

@article{roohani2024predicting,
  title={Predicting transcriptional outcomes of novel multigene perturbations with GEARS},
  author={Roohani, Yusuf and Huang, Kexin and Leskovec, Jure},
  journal={Nature Biotechnology},
  volume={42},
  number={6},
  pages={927--935},
  year={2024},
  publisher={Nature Publishing Group US New York}
}

@article{maathuis2010predicting,
  title={Predicting causal effects in large-scale systems from observational data},
  author={Maathuis, Marloes H and Colombo, Diego and Kalisch, Markus and B{\"u}hlmann, Peter},
  journal={Nature methods},
  volume={7},
  number={4},
  pages={247--248},
  year={2010},
  publisher={Nature Publishing Group US New York}
}

@inproceedings{schneidergenerative,
  title={Generative Intervention Models for Causal Perturbation Modeling},
  author={Schneider, Nora and Lorch, Lars and Kilbertus, Niki and Sch{\"o}lkopf, Bernhard and Krause, Andreas},
  booktitle={Forty-second International Conference on Machine Learning},
  year={2025}
}

@inproceedings{qu2025tabicl,
  title={TabICL: A Tabular Foundation Model for In-Context Learning on Large Data},
  author={Qu, Jingang and Holzm{\"u}ller, David and Varoquaux, Ga{\"e}l and Le Morvan, Marine},
  booktitle={International Conference on Machine Learning},
  pages={50817--50847},
  year={2025},
  organization={PMLR}
}

@article{grinsztajn2025tabpfn,
  title={Tabpfn-2.5: Advancing the state of the art in tabular foundation models},
  author={Grinsztajn, L{\'e}o and Fl{\"o}ge, Klemens and Key, Oscar and Birkel, Felix and Jund, Philipp and Roof, Brendan and J{\"a}ger, Benjamin and Safaric, Dominik and Alessi, Simone and Hayler, Adrian and others},
  journal={arXiv preprint arXiv:2511.08667},
  year={2025}
}

@article{srivatsan2020massively,
  title={Massively multiplex chemical transcriptomics at single-cell resolution},
  author={Srivatsan, Sanjay R and McFaline-Figueroa, Jos{\'e} L and Ramani, Vijay and Saunders, Lauren and Cao, Junyue and Packer, Jonathan and Pliner, Hannah A and Jackson, Dana L and Daza, Riza M and Christiansen, Lena and others},
  journal={Science},
  volume={367},
  number={6473},
  pages={45--51},
  year={2020},
  publisher={American Association for the Advancement of Science}
}

@article{aguirre2025gene,
  title={Gene regulatory network structure informs the distribution of perturbation effects},
  author={Aguirre, Matthew and Spence, Jeffrey P and Sella, Guy and Pritchard, Jonathan K},
  journal={PLOS Computational Biology},
  volume={21},
  number={9},
  pages={e1013387},
  year={2025},
  publisher={Public Library of Science San Francisco, CA USA}
}

@article{dibaeinia2020sergio,
  title={SERGIO: a single-cell expression simulator guided by gene regulatory networks},
  author={Dibaeinia, Payam and Sinha, Saurabh},
  journal={Cell systems},
  volume={11},
  number={3},
  pages={252--271},
  year={2020},
  publisher={Elsevier}
}

@article{sextro2026mappfn,
  title={MapPFN: Learning Causal Perturbation Maps in Context},
  author={Sextro, Marvin and K{\l}os, Weronika and Dernbach, Gabriel},
  journal={arXiv preprint arXiv:2601.21092},
  year={2026}
}

@inproceedings{dhirmeta,
  title={A Meta-Learning Approach to Bayesian Causal Discovery},
  author={Dhir, Anish and Ashman, Matthew and Requeima, James and van der Wilk, Mark},
  booktitle={The Thirteenth International Conference on Learning Representations},
  year={2025}
}

@article{guo2023causal,
  title={Causal de finetti: On the identification of invariant causal structure in exchangeable data},
  author={Guo, Siyuan and T{\'o}th, Viktor and Sch{\"o}lkopf, Bernhard and Husz{\'a}r, Ferenc},
  journal={Advances in Neural Information Processing Systems},
  volume={36},
  pages={36463--36475},
  year={2023}
}

@inproceedings{hagele2023bacadi,
  title={Bacadi: Bayesian causal discovery with unknown interventions},
  author={H{\"a}gele, Alexander and Rothfuss, Jonas and Lorch, Lars and Somnath, Vignesh Ram and Sch{\"o}lkopf, Bernhard and Krause, Andreas},
  booktitle={International Conference on Artificial Intelligence and Statistics},
  pages={1411--1436},
  year={2023},
  organization={PMLR}
}

@inproceedings{squires2020permutation,
  title={Permutation-based causal structure learning with unknown intervention targets},
  author={Squires, Chandler and Wang, Yuhao and Uhler, Caroline},
  booktitle={Conference on Uncertainty in Artificial Intelligence},
  pages={1039--1048},
  year={2020},
  organization={PMLR}
}

@article{mooij2020joint,
  title={Joint causal inference from multiple contexts},
  author={Mooij, Joris M and Magliacane, Sara and Claassen, Tom},
  journal={Journal of machine learning research},
  volume={21},
  number={99},
  pages={1--108},
  year={2020}
}

@book{spirtes2000causation,
  title={Causation, prediction, and search},
  author={Spirtes, Peter and Glymour, Clark N and Scheines, Richard},
  year={2000},
  publisher={MIT press}
}

@inproceedings{muller2025position,
  title={Position: The Future of Bayesian Prediction Is Prior-Fitted},
  author={M{\"u}ller, Samuel and Reuter, Arik and Hollmann, Noah and R{\"u}gamer, David and Hutter, Frank},
  booktitle={International Conference on Machine Learning},
  pages={81861--81875},
  year={2025},
  organization={PMLR}
}

@article{barabasi1999emergence,
  title={Emergence of scaling in random networks},
  author={Barab{\'a}si, Albert-L{\'a}szl{\'o} and Albert, R{\'e}ka},
  journal={science},
  volume={286},
  number={5439},
  pages={509--512},
  year={1999},
  publisher={American Association for the Advancement of Science}
}

@article{miladinovic2025silico,
  title={In-silico biological discovery with large perturbation models},
  author={Miladinovic, Djordje and H{\"o}ppe, Tobias and Chevalley, Mathieu and Georgiou, Andreas and Stuart, Lachlan and Mehrjou, Arash and Bantscheff, Marcus and Sch{\"o}lkopf, Bernhard and Schwab, Patrick},
  journal={arXiv preprint arXiv:2503.23535},
  year={2025}
}

@article{cui2023scGPT,
title={scGPT: Towards Building a Foundation Model for Single-Cell Multi-omics Using Generative AI},
author={Cui, Haotian and Wang, Chloe and Maan, Hassaan and Pang, Kuan and Luo, Fengning and Wang, Bo},
journal={bioRxiv},
year={2023},
publisher={Cold Spring Harbor Laboratory}
}

@article {Chen2023GenePT,
	author = {Chen, Yiqun T. and Zou, James},
	title = {GENEPT: A SIMPLE BUT HARD-TO-BEAT FOUNDATION MODEL FOR GENES AND CELLS BUILT FROM CHATGPT},
	elocation-id = {2023.10.16.562533},
	year = {2023},
	publisher = {Cold Spring Harbor Laboratory},
	journal = {bioRxiv}
}

@article{lotfollahi2019scgen,
  title={scGen predicts single-cell perturbation responses},
  author={Lotfollahi, Mohammad and Wolf, F Alexander and Theis, Fabian J},
  journal={Nature methods},
  volume={16},
  number={8},
  pages={715--721},
  year={2019},
  publisher={Nature Publishing Group US New York}
}

@article{piran2024disentanglement,
  title={Disentanglement of single-cell data with biolord},
  author={Piran, Zoe and Cohen, Niv and Hoshen, Yedid and Nitzan, Mor},
  journal={Nature Biotechnology},
  volume={42},
  number={11},
  pages={1678--1683},
  year={2024},
  publisher={Nature Publishing Group US New York}
}

@article{bereket2023modelling,
  title={Modelling cellular perturbations with the sparse additive mechanism shift variational autoencoder},
  author={Bereket, Michael and Karaletsos, Theofanis},
  journal={Advances in Neural Information Processing Systems},
  volume={36},
  pages={1--12},
  year={2023}
}

@article{bunne2022supervised,
  title={Supervised training of conditional monge maps},
  author={Bunne, Charlotte and Krause, Andreas and Cuturi, Marco},
  journal={Advances in Neural Information Processing Systems},
  volume={35},
  pages={6859--6872},
  year={2022}
}

@article{liu2026learning,
  title={Learning genetic perturbation effects with variational causal inference},
  author={Liu, Emily and Zhang, Jiaqi and Uhler, Caroline},
  journal={PLOS Computational Biology},
  volume={22},
  number={2},
  pages={e1013194},
  year={2026},
  publisher={Public Library of Science San Francisco, CA USA}
}

@article{qu2026tabiclv2,
  title={TabICLv2: A better, faster, scalable, and open tabular foundation model},
  author={Qu, Jingang and Holzm{\"u}ller, David and Varoquaux, Ga{\"e}l and Morvan, Marine Le},
  journal={arXiv preprint arXiv:2602.11139},
  year={2026}
}

@article{balazadeh2025causalpfn,
  title={CausalPFN: Amortized causal effect estimation via in-context learning},
  author={Balazadeh, Vahid and Kamkari, Hamidreza and Thomas, Valentin and Li, Benson and Ma, Junwei and Cresswell, Jesse C and Krishnan, Rahul G},
  journal={arXiv preprint arXiv:2506.07918},
  year={2025}
}

@inproceedings{
dhir2026estimating,
title={Estimating Interventional Distributions with Uncertain Causal Graphs through Meta-Learning},
author={Anish Dhir and Cristiana Diaconu and Valentinian Mihai Lungu and James Requeima and Richard E. Turner and Mark van der Wilk},
booktitle={The Thirty-ninth Annual Conference on Neural Information Processing Systems},
year={2026}
}

@article{ma2025foundation,
  title={Foundation models for causal inference via prior-data fitted networks},
  author={Ma, Yuchen and Frauen, Dennis and Javurek, Emil and Feuerriegel, Stefan},
  journal={arXiv preprint arXiv:2506.10914},
  year={2025}
}

@article{theodoris2023transfer,
  title={Transfer learning enables predictions in network biology},
  author={Theodoris, Christina V and Xiao, Ling and Chopra, Anant and Chaffin, Mark D and Al Sayed, Zeina R and Hill, Matthew C and Mantineo, Helene and Brydon, Elizabeth M and Zeng, Zexian and Liu, X Shirley and others},
  journal={Nature},
  volume={618},
  number={7965},
  pages={616--624},
  year={2023},
  publisher={Nature Publishing Group UK London}
}

@article{dong2026stack,
  title={Stack: In-Context Learning of Single-Cell Biology},
  author={Dong, Mingze and Adduri, Abhinav and Gautam, Dhruv and Carpenter, Christopher and Shah, Rohan and Ricci-Tam, Chiara and Kluger, Yuval and Burke, Dave P and Roohani, Yusuf H},
  journal={bioRxiv},
  year={2026}
}

@article{zheng2018dags,
  title={Dags with no tears: Continuous optimization for structure learning},
  author={Zheng, Xun and Aragam, Bryon and Ravikumar, Pradeep K and Xing, Eric P},
  journal={Advances in neural information processing systems},
  volume={31},
  year={2018}
}

@article{kunzel2019metalearners,
  title={Metalearners for estimating heterogeneous treatment effects using machine learning},
  author={K{\"u}nzel, S{\"o}ren R and Sekhon, Jasjeet S and Bickel, Peter J and Yu, Bin},
  journal={Proceedings of the national academy of sciences},
  volume={116},
  number={10},
  pages={4156--4165},
  year={2019},
  publisher={National Academy of Sciences}
}

@article{nourisa2025genernib,
  title={geneRNIB: a living benchmark for gene regulatory network inference},
  author={Nourisa, Jalil and Passemiers, Antoine and Kalfon, Jeremie and Stock, Marco and Zeller-Plumhoff, Berit and Cannoodt, Robrecht and Arnold, Christian and Tong, Alexander and Hartford, Jason and Netea, Mihai G and others},
  journal={bioRxiv},
  pages={2025--02},
  year={2025},
  publisher={Cold Spring Harbor Laboratory}
}

@article{kim2015ppcor,
  title={ppcor: an R package for a fast calculation to semi-partial correlation coefficients},
  author={Kim, Seongho},
  journal={Communications for statistical applications and methods},
  volume={22},
  number={6},
  pages={665},
  year={2015}
}

@article{moerman2019grnboost2,
  title={GRNBoost2 and Arboreto: efficient and scalable inference of gene regulatory networks},
  author={Moerman, Thomas and Aibar Santos, Sara and Bravo Gonz{\'a}lez-Blas, Carmen and Simm, Jaak and Moreau, Yves and Aerts, Jan and Aerts, Stein},
  journal={Bioinformatics},
  volume={35},
  number={12},
  pages={2159--2161},
  year={2019},
  publisher={Oxford University Press}
}

@article{passemiers2022fast,
  title={Fast and accurate inference of gene regulatory networks through robust precision matrix estimation},
  author={Passemiers, Antoine and Moreau, Yves and Raimondi, Daniele},
  journal={Bioinformatics},
  volume={38},
  number={10},
  pages={2802--2809},
  year={2022},
  publisher={Oxford University Press}
}

@article{aibar2017scenic,
  title={SCENIC: single-cell regulatory network inference and clustering},
  author={Aibar, Sara and Gonz{\'a}lez-Blas, Carmen Bravo and Moerman, Thomas and Huynh-Thu, V{\^a}n Anh and Imrichova, Hana and Hulselmans, Gert and Rambow, Florian and Marine, Jean-Christophe and Geurts, Pierre and Aerts, Jan and others},
  journal={Nature methods},
  volume={14},
  number={11},
  pages={1083--1086},
  year={2017},
  publisher={Nature Publishing Group US New York}
}


\newpage
\appendix

\section{Details on Synthetic Data Generation}
\label{app:synthetic}

PerturbPFN is pretrained on episodes sampled from a hierarchical synthetic SCM prior. Each episode contains a task-level regulatory system, a collection of drug- and dose-dependent interventions, and context and query responses generated from the same latent system. The graph, targets, strengths, latent states, and observed responses are retained as synthetic supervision, but the model receives only regime descriptors and response observations as inputs.

\paragraph{Episode-level graph and mechanisms.}

\begin{table}[t]
\centering
\caption{Main hyperparameters of the synthetic graph prior.}
\label{tab:synthetic-grn-prior}
\small
\begin{tabular}{lll}
\toprule
Symbol & Description & Range / value \\
\midrule
\(p\) & Number of genes & \([40,60]\) \\
\(K\) & Number of gene modules & \([4,8]\) \\
\(\alpha_0\) & Base edge logit & \(-2.5\) \\
\(\sigma_s\) & Source-scale log-normal scale & \(0.9\) \\
\(\alpha_t\) & Target-scale Gamma shape & \(2.0\) \\
\(\mu_B,\sigma_B\) & Module-pair logit parameters & \(-0.2,\,0.35\) \\
\(\beta_{\mathrm{within}}\) & Within-module logit boost & \(1.35\) \\
\(\lambda_{\mathrm{dist}}\) & Topological-distance penalty & \(1.5\) \\
\(\mu_w,\sigma_w\) & Edge-weight log-normal parameters & \(-0.2,\,0.45\) \\
\(p_+\) & Positive-edge probability & \(0.7\) \\
\bottomrule
\end{tabular}
\end{table}

The graph prior is designed to produce sparse, directed, degree-heterogeneous, and modular GRN-like systems \citep{aguirre2025gene,bartlett2026inferring}. For each episode, genes are assigned to balanced modules and placed in a random topological order. Only forward edges \(u\to v\), with \(u<v\), are allowed. Their probabilities are
\[
A_{uv}\sim\mathrm{Bernoulli}\!\left(\sigma(\ell_{uv})\right),
\qquad
\ell_{uv}
=
\alpha_0+\log s_u+0.35\log t_v
+B_{c_u,c_v}
-\lambda_{\mathrm{dist}}\frac{v-u}{p-1},
\]
where \(s_u\) and \(t_v\) are source- and target-specific activity factors, \(c_u\) denotes the module of gene \(u\), and \(B\) is a sampled module-pair affinity matrix with an additional within-module boost. Existing edges receive signed log-normal weights. This construction guarantees acyclicity while inducing modular and heterogeneous connectivity.

Each node is assigned an independently sampled two-layer nonlinear mechanism, a basal expression level, and an exogenous-noise scale. In contrast to the dose-response model below, graph propagation is implemented by a \(\tanh\) MLP rather than by Hill functions.

\paragraph{Perturbation descriptors, targets, and strengths.}

\begin{table}[t]
\centering
\caption{Main perturbation and observation settings used for the final checkpoint.}
\label{tab:synthetic-perturbation-prior}
\small
\begin{tabular}{lll}
\toprule
Symbol & Description & Range / value \\
\midrule
\(F\) & Number of drug families & \([3,6]\) \\
\(D\) & Number of synthetic drugs & \([12,24]\) \\
\(d_{\mathrm{emb}}\) & Drug embedding dimension & \(8\) \\
\(M_{\max}\) & Preferred modules per family & \(1\)--\(2\) \\
\(p_{\mathrm{stop}}\) & Target-count stop probability & \(1/3\) \\
\(\mathbb{E}[|I|]\) & Expected target count & \(\approx 3\) \\
\(a_{\min},a_{\max}\) & Dose range & \([0.02,10.0]\) \\
\(L\) & Held-out-drug dose levels & \(4\) \\
\(n_{\min},n_{\max}\) & Family-level Hill range & \([2.0,5.0]\) \\
\(\mu_{E},\sigma_{E}\) & \(E^{\max}\) log-normal parameters & \(0.2,\,0.1\) \\
\(C\) & Simulated cells per condition & \(128\) \\
\(n_{\mathrm{ctx}}/n_{\mathrm{q}}\) & Context / query entries & \(20/4\) \\
\(\sigma_{\mathrm{obs}}\) & Observation log-noise scale & \(0.001\) \\
\(p_{0,\max}\) & Maximum zero probability & \(0.001\) \\
\bottomrule
\end{tabular}
\end{table}

Synthetic drugs are organized into latent families. Each family has high affinity for one or two gene modules and lower affinity for the remaining modules. Drugs inherit a normalized family embedding with drug-specific noise, making nearby descriptors statistically more likely to induce related interventions. The descriptor for drug \(d\) at dose \(a\) is
\[
\bm{\gamma}_{d,a}
=
\left[
\frac{a}{1+a},
\log(1+a),
\mathbb{I}\{\mathrm{control}\},
\mathbf e_d
\right],
\]
where \(\mathbf e_d\in\mathbb{R}^8\) is the drug embedding; controls use a zero embedding and zero dose.

For drug \(d\), the target propensity of gene \(i\) is
\[
\pi_{d,i}
\propto
A_{f_d,c_i}\,q_i\,
\exp\!\left(\sigma_{\mathrm{drug}}\xi_{d,i}\right),
\qquad
\xi_{d,i}\sim\mathcal N(0,1),
\]
where \(A_{f_d,c_i}\) is the family--module affinity and \(q_i\) is a weak gene-specific targetability factor derived from graph degree and module frequency. The target count follows a truncated geometric distribution beginning at one with \(p_{\mathrm{stop}}=1/3\), giving approximately three targets per intervention. Targets are sampled without replacement according to \(\pi_{d,i}\).

For target \(i\), the signed dose-dependent strength is
\[
\psi_{d,a,i}
=
s_{d,i}E^{\max}_{d,i}
\frac{a^{n_{d,i}}}
     {(EC50_{d,i})^{n_{d,i}}+a^{n_{d,i}}},
\]
where \(s_{d,i}\in\{-1,1\}\). The \(E^{\max}\), \(EC50\), Hill coefficient, and sign distributions share family--module parameters with drug- and gene-specific variation. Non-target genes and control conditions have zero strength.

\paragraph{Latent rollout and observation model.}
For each cell, genes are generated in topological order. Let \(\mathbf w_{:v}\) denote incoming edge weights for gene \(v\). Its latent log-state is
\[
\begin{aligned}
\mathbf h_v
&=
\tanh\!\left(
W_{1,v}^{\top}
\bigl(\mathbf z\odot A_{:v}\odot\mathbf w_{:v}\bigr)
+\mathbf b_{1,v}
\right),\\
z_v
&=
b_v+\mathbf w_{2,v}^{\top}\mathbf h_v+b_{2,v}
+I_v\psi_v+\sigma_v\epsilon_v .
\end{aligned}
\]
Thus, the intervention enters once as an additive shift to the targeted local mechanism and propagates to descendants through the graph.

The simulator maps latent states to expression using a zero-inflated log-normal observation model inspired by single-cell simulators such as SERGIO \citep{dibaeinia2020sergio}. The implementation supports expression-dependent zero inflation and heteroscedastic log-normal noise. For the final checkpoint, however, we use a low-corruption configuration with log-noise scale \(0.001\) and zero probability at most \(0.001\). This preserves latent intervention and graph signals while retaining the same observation-model parameterization as the decoder. Each condition contains \(128\) simulated cells, and their mean expression is supplied as the condition-level response.

\paragraph{Highest-dose episodic protocol.}
Training episodes follow the highest-dose holdout structure used in the real-data evaluation. One synthetic drug is selected as the held-out drug and evaluated on a four-level geometric dose ladder. The context contains a control, its observed lower-dose conditions, and conditions from other drugs; the query corresponds to its held-out highest dose. All conditions within an episode reuse the same exogenous base-noise matrix for latent rollout, producing paired responses whose differences primarily reflect interventions rather than independent cell noise \citep{sextro2026mappfn}. The model does not observe target propensities or target identities as inputs; these quantities, together with the graph and strengths, are used only as synthetic supervision.

\section{Model Architecture and Training}
\label{app:architecture}

Sections~\ref{sec:architecture} and~\ref{sec:training-inference} describe the model computation and training objective. Here, we report the exact architecture dimensions, optimization settings, and checkpoint-selection procedure used in the experiments.

\begin{table}[h]
\centering
\caption{Architecture settings of the final PerturbPFN model.}
\label{tab:model-architecture}
\small
\begin{tabular}{lll}
\toprule
Component & Setting & Value \\
\midrule
Input & Maximum nodes / metadata dim. & \(60 / 11\) \\
Shared encoder & Hidden dim. / heads / blocks & \(192 / 6 / 5\) \\
Shared encoder & Feedforward dimension & \(640\) \\
Context memory & Token dim. / attention heads & \(192 / 4\) \\
Structure encoder & Hidden dim. / heads / blocks & \(256 / 8 / 4\) \\
Structure encoder & Feedforward dimension & \(1024\) \\
Structure evidence & Pooling / hidden dimension & mean--max / \(64\) \\
Structure head & Edge scorer & biaffine + contextual evidence \\
Intervention heads & MLP hidden dimension & \(320\) \\
Target head & Straight-through top-\(k\) & \(3\) \\
Strength head & Number of bins / support & \(81 / [-2,2]\) \\
SCM decoder & Hidden dim. / layers & \(64 / 2\) \\
SCM decoder & Activation / log-variance range & \(\tanh / [-6,3]\) \\
Parameters & Total parameter count & \(14.98\)M \\
\bottomrule
\end{tabular}
\end{table}

The structure encoder uses pairwise context representations with mean--max aggregation. Its edge scorer combines a biaffine node-pair score with learned context-pair and intervention--response features. Context intervention memory uses predicted context targets and strengths during joint training and inference; synthetic intervention labels are used only for supervision and structure pretraining. The strength estimate is the expectation of the predicted bar distribution. The decoder uses separate parent slots and propagates latent log-states in topological order.

\begin{table}[h]
\centering
\caption{Training settings and checkpoint-selection procedure.}
\label{tab:training-hparams}
\small
\begin{tabular}{lll}
\toprule
Category & Setting & Value \\
\midrule
Reproducibility & Random seed & \(17\) \\
Synthetic protocol & Dose levels & \(4\) \\
Synthetic batch & Batch size & \(8\) \\
Synthetic batch & Context / query regimes & \(20 / 4\) \\
Optimizer & Optimizer & AdamW \\
Optimizer & Learning rate / weight decay & \(3\times10^{-4} / 10^{-4}\) \\
Optimizer & Gradient clipping & \(1.0\) \\
Loss weights & Graph / target / strength & \(1 / 2 / 2\) \\
Context supervision & Target / strength weights & \(1 / 1\) \\
Target loss & Positive weight / focal exponent & \(10 / 1\) \\
Strength loss & Non-target penalty & \(0.05\) \\
Graph loss & Density regularizer & \(1.0\) \\
Acyclicity & Initial / maximum penalty & \(1 / 100\) \\
Curriculum & Structure pretraining & \(2500\) steps \\
Curriculum & Graph-GT joint training & \(500\) steps \\
Continuation & Soft-graph training & \(500\) steps \\
Training run & Total executed steps & \(3500\) \\
Model selection & Selected checkpoint & step \(3100\) \\
\bottomrule
\end{tabular}
\end{table}

During structure pretraining, only the structure encoder and structure head are optimized. The subsequent graph-GT stage jointly optimizes the complete model while using the ground-truth synthetic graph for decoder rollout. The run was continued for \(500\) soft-graph steps to compare later checkpoints, and development-set selection chose step \(3100\). All reported downstream results use this checkpoint without parameter updates.

\paragraph{Compute resources.}
The selected model was trained on a single NVIDIA RTX 4090 GPU with \(24\)GB of memory. The complete \(3500\)-step checkpoint-selection run took approximately \(1.44\) GPU-hours with batch size \(8\). The selected step-\(3100\) checkpoint contains \(14.98\)M parameters and is shared across all reported evaluation tasks.

\section{Details on In-Prior Evaluation}
\label{app:in-prior}

\paragraph{Graph recovery.}
Graph AUROC and AUPRC are computed from the predicted probabilities of all non-diagonal directed edges. For projected F1, we retain the highest-scoring edges up to the ground-truth edge count, skipping edges that would create a directed cycle, and compare the resulting DAG with the ground-truth graph.

\paragraph{Target recovery.}
Target AUROC and AUPRC are computed from the node-wise target probabilities. The reported target F1 thresholds these probabilities at \(0.5\); it does not use the straight-through top-\(k\) gate employed during SCM rollout.

\paragraph{Strength recovery.}
Strength calibration is measured on ground-truth target nodes using
\[
R_{\psi}
=
\frac{
\mathbb{E}_{I_{j,v}=1}\!\left[|\hat{\psi}_{j,v}|\right]
}{
\mathbb{E}_{I_{j,v}=1}\!\left[|\psi_{j,v}|\right]
}.
\]
A value near \(1\) indicates that the predicted intervention strengths recover the correct overall scale.

\paragraph{Response recovery.}
Delta cosine is the cosine similarity between the predicted and true mean response shifts relative to control, after centering both shifts across genes. Observation NLL is the average negative log-likelihood of the query responses under the predicted zero-inflated log-normal distribution.

\section{Details on Effect Prediction}
\label{app:effect-prediction}

This section provides additional details on the real-data effect-prediction benchmark, including the dataset preprocessing, highest-dosage holdout protocol, evaluation metrics, and baseline implementations.

\subsection{Dataset}
We evaluate on the Sci-Plex single-cell drug perturbation dataset of \citet{srivatsan2020massively}, following the preprocessed epigenetic-regulation benchmark used by \citet{schneidergenerative}. The benchmark contains three cell lines, A549, K562, and MCF7, and four small-molecule perturbations, belinostat, dacinostat, givinostat, and quisinostat 2HCl, giving \(3\times4=12\) held-out protocols. Each protocol is defined by one cell line--drug pair.

We use the same preprocessing convention as the GIM benchmark. Cells are normalized to the median count of control cells, filtered, log-transformed, and restricted to a protocol-specific set of \(50\) marker genes selected using only the training conditions. Perturbation descriptors encode drug identity and dose information. Control cells are included as the observational regime.

\subsection{Highest-Dosage Holdout Protocol}
For each cell line--drug protocol, the highest dose of the held-out drug is removed from the context and used as the query condition. The context set contains control cells, lower doses of the held-out drug, and all available doses of the remaining drugs in the same cell line. The model must predict the distribution of expression vectors under the held-out highest dose.

This protocol tests dose extrapolation under unknown perturbation targets. The held-out drug is not completely unseen, since lower-dose responses are available, but the target and mechanism at the highest dose are not provided to the model. All methods are evaluated on the same held-out cells and with the same metric implementation.

\subsection{Metrics}

Let \(X_{\mathrm{ctrl}}\), \(X_{\mathrm{int}}\), and
\(\hat{X}_{\mathrm{int}}\) denote control cells, true held-out
perturbed cells, and predicted perturbed cells, respectively. We
evaluate three complementary metrics.

\paragraph{Mean distance.}
We measure mean-response error using
\[
\mathrm{MD}
=
\left\|
\mathbb{E}[\hat{X}_{\mathrm{int}}]
-
\mathbb{E}[X_{\mathrm{int}}]
\right\|_2 ,
\]
where lower values indicate more accurate prediction of the average
response.

\paragraph{Pearson correlation.}
We compute
\[
r
=
\mathrm{corr}
\left(
\mathbb{E}[\hat{X}_{\mathrm{int}}],
\mathbb{E}[X_{\mathrm{int}}]
\right),
\]
which measures gene-wise agreement between the predicted and true
mean-response patterns. Higher values are better.

\paragraph{Wasserstein distance.}
We approximate the \(2\)-Wasserstein distance using the square root
of a debiased entropic Sinkhorn transport cost with squared-Euclidean
ground cost:
\[
\widehat W_2(X,Y)
=
\left[
C_\epsilon(X,Y)
-\frac{1}{2}C_\epsilon(X,X)
-\frac{1}{2}C_\epsilon(Y,Y)
\right]_{+}^{1/2},
\]
where \(C_\epsilon\) is the transport cost under the entropically
regularized Sinkhorn coupling. We use \(\epsilon=0.1\), at most
\(200\) iterations, and tolerance \(10^{-6}\). Predicted and true
distributions are evaluated using equally sized, deterministically
sampled cell sets. Lower values are better.

\paragraph{Normalized multi-metric score.}
For Table~\ref{tab:runtime-per-protocol}, each metric is oriented so
that higher is better and min--max normalized within each protocol:
\[
s_{m,t,k}
=
\frac{z_{m,t,k}-z^{\min}_{t,k}}
     {z^{\max}_{t,k}-z^{\min}_{t,k}},
\]
where \(z=-\mathrm{MD}\), \(z=r\), or \(z=-\widehat W_2\).
We first average each normalized metric across the \(12\) protocols
and then average the three metric scores to obtain the overall score
and rank. These normalized scores depend on the included comparison
set.

\subsection{Baselines}
We compare against structured causal, distributional, supervised, and heuristic baselines. All baselines are evaluated on the same held-out protocols, and all reported metrics are recomputed using the same metric code as PerturbPFN whenever prediction arrays are available.

\paragraph{GIM.}
GIM \citep{schneidergenerative} is the closest structured causal baseline. We use the authors' implementation and Sci-Plex preprocessing. GIM is fit separately for each held-out protocol by instance-specific optimization. We use the Sci-Plex configuration from the GIM codebase: a nonlinear zero-inflated log-normal inference model, unknown intervention targets, Adam optimization with step size \(10^{-3}\), graph prior and intervention sparsity penalties, and augmented-Lagrangian acyclicity optimization. The main GIM runs use \(100{,}000\) optimization steps following the original setting.

\paragraph{CondOT.}
CondOT \citep{bunne2022supervised} is a conditional optimal-transport baseline that learns a perturbation-conditioned map from control cells to treated-cell distributions. We use the PICNN-based implementation from the GIM benchmark, with perturbation descriptors encoding drug identity and dose. The two transport potentials each contain four hidden layers with \(128\) units and are optimized for \(100{,}000\) alternating steps using Adam with learning rate \(10^{-4}\) and coefficients \((\beta_1,\beta_2)=(0.5,0.9)\). CondOT is fitted separately for each held-out protocol using only the corresponding training conditions. Predictions are generated by transporting sampled control cells under the descriptor of the held-out highest-dose condition.

\paragraph{MLP.}
The MLP baseline follows the shift-regression baseline from the GIM codebase. It learns a mapping from perturbation descriptors to mean expression shifts relative to the control distribution. At prediction time, the estimated shift for the held-out descriptor is added to sampled control cells to form predicted perturbed cells. We use a two-layer MLP regressor with hidden size \(100\), learning rate \(10^{-3}\), and maximum \(20{,}000\) iterations.

\paragraph{RF X-Learner.}
RF X-Learner is a supervised treatment-effect baseline based on the X-learner meta-learning strategy \citep{kunzel2019metalearners}. The model first fits a global response model \(\mu(\bm{\gamma})\), then constructs imputed treatment effects for treated and control samples and fits two effect models \(\tau_1(\bm{\gamma})\) and \(\tau_0(\bm{\gamma})\). At prediction time, the final effect is a global constant-weight mixture of \(\tau_0\) and \(\tau_1\), added to sampled control cells to generate the predicted perturbed distribution. The backbone is a multi-output \texttt{RandomForestRegressor} with \(100\) trees, no maximum depth, full training data, and no downsampling; targets are z-scored during fitting and transformed back to the original expression scale for evaluation.

\paragraph{Naive and Observational.}
The Naive baseline predicts the held-out highest-dose response using the nearest observed lower dose of the same drug in the same cell line. The Observational baseline ignores the perturbation and predicts the control distribution. These training-free methods serve as simple reference baselines for
dose extrapolation and perturbation-response prediction.

\section{Details on Regulatory Structure Discovery}
\label{app:regulatory-structure}

\subsection{GeneRNIB 300BCG Protocol}

We evaluate regulatory-structure transfer using the official GeneRNIB 300BCG benchmark \citep{nourisa2025genernib}. PerturbPFN receives only the benchmark inference matrix, which contains \(380\) log-normalized expression profiles. The held-out evaluation data and biological reference networks are not provided to the model. Numerical metadata are standardized, categorical metadata are encoded into the model's \(11\)-dimensional regime descriptor, and episode sampling is balanced across the available cell-type labels.

The same pretrained checkpoint used in the other experiments is applied without gradient updates. Because PerturbPFN predicts an episode-level graph, we construct stochastic episodes from the inference profiles and average its directed edge probabilities across episodes. GeneRNIB requires a weighted TF--gene network with at most \(50{,}000\) directed links.

\subsection{Panel Construction and Network Aggregation}

PerturbPFN supports at most \(60\) nodes per forward pass, whereas the GeneRNIB expression matrix contains substantially more genes. We therefore apply the model to overlapping TF--target panels and aggregate their edge scores. Candidate regulators are restricted to the official GeneRNIB TF list. We select the \(100\) overlapping TFs with the highest expression variance as panel seeds and select \(1{,}500\) candidate target genes, prioritizing genes marked as highly variable and then ranking by expression variance.

The TF seeds are divided into groups of \(20\), and the target genes into groups of \(40\). Each pair of TF and target groups forms a panel containing at most \(60\) genes, resulting in \(190\) panels. TFs that also occur in a target group remain eligible as regulators. Table~\ref{tab:genernib-inference-settings} summarizes the inference settings.

\begin{table}[t]
\centering
\caption{PerturbPFN inference settings for GeneRNIB 300BCG.}
\label{tab:genernib-inference-settings}
\small
\begin{tabular}{ll}
\toprule
Setting & Value \\
\midrule
Input expression layer & Log-normalized \\
Input profiles & \(380\) \\
Maximum nodes per panel & \(60\) \\
High-variance TF seeds & \(100\) \\
Candidate target genes & \(1{,}500\) \\
TF / target genes per panel & \(20 / 40\) \\
Number of panels & \(190\) \\
Episodes per panel & \(32\) \\
Context / query profiles & \(64 / 16\) \\
Maximum output links & \(50{,}000\) \\
\bottomrule
\end{tabular}
\end{table}

For each panel, we sample \(32\) episodes, each containing \(64\) context and \(16\) query profiles. The graph head is episode-level and is inferred from the context data; its edge probabilities are averaged across episodes. For every ordered TF--gene pair \(u\to v\), \(u\neq v\), the resulting score is
\[
s_{uv}
=
\frac{1}{32}
\sum_{e=1}^{32}
P^{(e)}_{uv}.
\]
When an edge occurs in multiple overlapping panels, we retain its maximum score. All aggregated edges are then sorted by score, and the highest-scoring \(50{,}000\) links are exported in the official GeneRNIB format.

Although PerturbPFN projects its graph to a DAG for SCM rollout, the GeneRNIB submission uses the continuous, unthresholded edge probabilities rather than the projected adjacency. This preserves the edge ranking required by GeneRNIB and does not impose acyclicity on the exported GRN, which may contain regulatory cycles.

\subsection{Official Evaluation and Baselines}

We evaluate the exported network using the official GeneRNIB \texttt{all\_metrics} pipeline with ridge regression for the regression-based metrics. The scorer uses the held-out 300BCG expression data, the benchmark regulator consensus, and TF-binding references from UniBind, ChIP-Atlas, and ReMap. None of these evaluation resources are accessed during PerturbPFN inference.

Gene-set F1 (GS-F1) measures agreement between pathways enriched among predicted TF targets and pathways supported by the held-out expression data. TF-binding F1 (TFB-F1) evaluates agreement with the external TF-binding references. Regression precision and recall (R-Precision and R-Recall) assess whether regulators selected from the inferred network predict held-out target-gene expression under conservative and broader regulator budgets. The virtual-cell score (VC) evaluates perturbation prediction using a model constrained by the inferred network. The official pipeline additionally reports replicate consistency, which is omitted from the main table for space. Higher values are better for all metrics.

We compare against Pearson correlation, partial Pearson correlation (PPCOR), GRNBoost2, PORTIA, and SCENIC. Baseline results are taken from the official GeneRNIB score archive for the same 300BCG dataset and \(50{,}000\)-link budget. Thus, every method is compared using the same held-out evaluation data, reference resources, and metric implementations.



\end{document}